\documentclass[conference]{IEEEtran}
\IEEEoverridecommandlockouts

\usepackage{cite}
\usepackage{amsmath,amssymb,amsfonts}
\usepackage{algorithmic}
\usepackage{graphicx}
\usepackage{textcomp}
\usepackage{xcolor}
\usepackage{subfigure}
\usepackage{amsthm}
\usepackage{caption}
\usepackage{subcaption}
\usepackage{bm}
\usepackage[ruled,longend,linesnumbered]{algorithm2e}
\usepackage{booktabs}
\usepackage{tabularx}

\theoremstyle{plain}
\newtheorem{definition}{Definition}
\usepackage[colorlinks,
            linkcolor=red,
            anchorcolor=blue,
            citecolor=green
            ]{hyperref}

\def\BibTeX{{\rm B\kern-.05em{\sc i\kern-.025em b}\kern-.08em
    T\kern-.1667em\lower.7ex\hbox{E}\kern-.125emX}}
\begin{document}

\title{Accelerating Outlier-robust Rotation Estimation by Stereographic Projection

\thanks{Taosi Xu, Yinlong Liu, and Zhi-Xin Yang are with the University of Macau, Macau, China, Email:\{taosixu, yinlongliu, zxyang\}@umac.mo.

*Zhi-Xin Yang is the corresponding author.

Xianbo Wang is with the Hainan Institute of Zhejiang University, Hainan, China, xbwang@zju.edu.cn}
}

\author{\IEEEauthorblockN{Taosi Xu, Yinlong Liu, Xianbo Wang, Zhi-Xin Yang*}}



\maketitle
\begin{abstract}
Rotation estimation plays a fundamental role in many computer vision and robot tasks. However, efficiently estimating rotation in large inputs containing numerous outliers (i.e., mismatches) and noise is a recognized challenge. Many robust rotation estimation methods have been designed to address this challenge. Unfortunately, existing methods are often inapplicable due to their long computation time and the risk of local optima. In this paper, we propose an efficient and robust rotation estimation method. 
Specifically, our method first investigates geometric constraints involving only the rotation axis. Then, it uses stereographic projection and spatial voting techniques to identify the rotation axis and angle. Furthermore, our method efficiently obtains the optimal rotation estimation and can estimate multiple rotations simultaneously. To verify the feasibility of our method, we conduct comparative experiments using both synthetic and real-world data. The results show that, with GPU assistance, our method can solve large-scale ($10^6$ points) and severely corrupted (90\% outlier rate) rotation estimation problems within 0.07 seconds, with an angular error of only 0.01 degrees, which is superior to existing methods in terms of accuracy and efficiency.
\end{abstract}
\begin{figure}[t]
\centering
\subfigure[]{\includegraphics[width=0.4\linewidth]{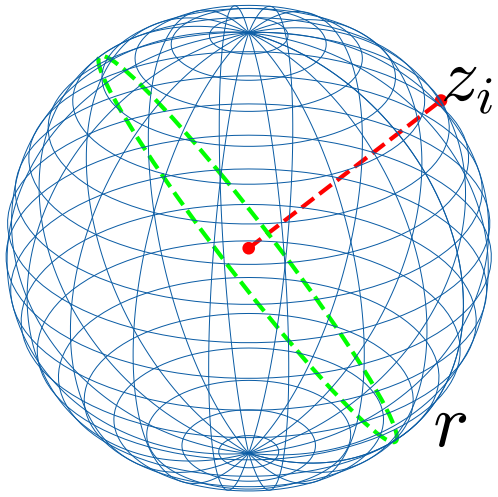}}
\subfigure[]{\includegraphics[width=0.44\linewidth]{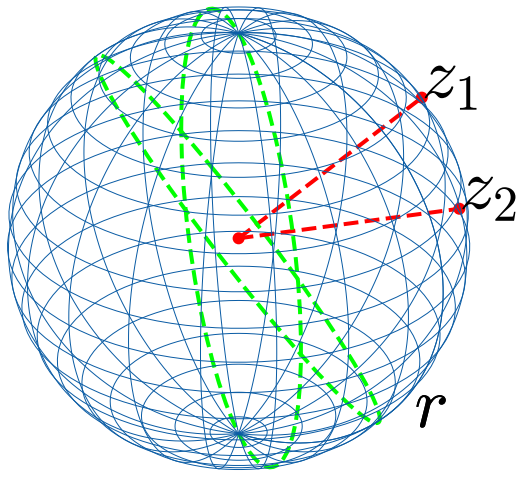}}
\\
\subfigure[]{\includegraphics[width=0.4\linewidth]{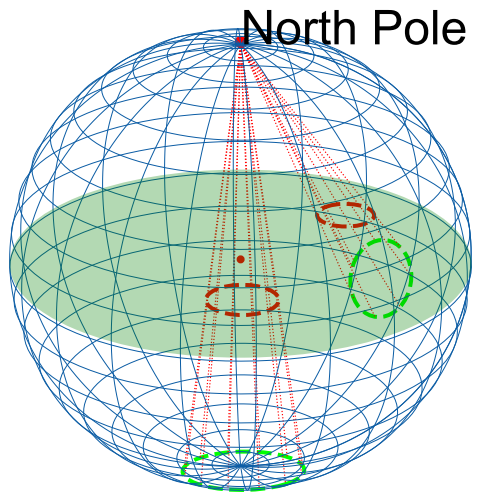}}
\subfigure[]{\includegraphics[width=0.4\linewidth]{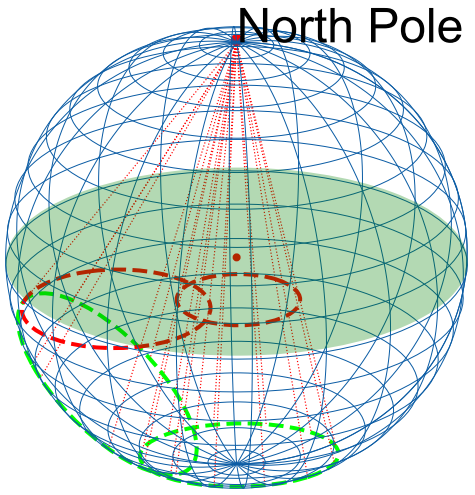}}
\caption{ (a) Given the $i$-th \(\bm{\mathit{z}}_{i}\), all possible rotation axes are in a circle (a green dashed circle). (b) Given \(\bm{\mathit{z}}_{1}\) and \(\bm{\mathit{z}}_{2}\), all possible rotation axes should be in the intersections of two corresponding circles. (c) The stereographic projection is circle-preserving, and a 3D circle on the sphere is into a 2D circle when projected onto the plane. (d) The intersections of 3D circles are projected to the intersections of 2D circles in the plane by stereographic projection.}
\label{fig2}
\end{figure}
\section{Introduction}
Rotation estimation is a significant problem in computer vision and robotics~\cite{bustos2016fast}. In this problem, two sets of 3D points are given, and the task is to optimally align these two sets by estimating the best rotation between them~\cite{huang2021comprehensive}. As a core part of rigid pose estimation, rotation estimation plays a crucial role in many intelligent applications~\cite{szeliski2022computer}, including robot perception~\cite{bernreiter2021phaser}, image stitching~\cite{parra2015guaranteed}, computer graphics~\cite{maron2016point}, and satellite attitude determination~\cite{chng2023rosia}. All of these applications rely on accurate and reliable rotation estimation solutions. Additionally, it is often necessary to estimate multiple rotations~\cite{farina2023quantum}, not just the optimal one. For instance, multi-model rotation estimation is crucial in areas like autonomous driving~\cite{jin2024multi}.

Mathematically, given point sets $\mathcal{Q} = \left \{ \bm{\mathit{q}}_{1},...,\bm{\mathit{q}}_{n} \right \} \subset \mathbb{R}^{3}$ and $\mathcal{P} = \left \{ \bm{\mathit{p}}_{1},...,\bm{\mathit{p}}_{n} \right \} \subset \mathbb{R}^{3}$, let $\mathcal{C}$ be a subset of $\left [ n \right ] \times \left [ n \right ] := \left \{ 1,...,n \right \} \times \left \{ 1,...,n \right \}$, called the inlier correspondence set. Assume that
\begin{equation}
\bm{\mathit{q}}_{i} = \mathbf{R}\bm{\mathit{p}}_{i} + \epsilon_{i}, \qquad  i \in \mathcal{C} \label{(1)}
\end{equation}
where $\epsilon_{i}$ is noise, $\mathbf{R}$ is a rotation, and $\left ( \bm{\mathit{q}}_{i}, \bm{\mathit{p}}_{i} \right )$ is called an \textit{inlier}. If the equation is not true, then $\left ( \bm{\mathit{q}}_{i}, \bm{\mathit{p}}_{i} \right )$ is called an \textit{outlier}~\cite{peng2022arcs}.
If we seek a single optimal rotation, the goal of Equation \ref{(1)} is to simultaneously estimate the global optimal 3D rotation $\mathbf{R}$ and the inlier correspondence set $\mathcal{C}$ from point sets $\mathcal{Q}$ and $\mathcal{P}$. If there are multiple rotations to be solved, the goal of Equation \ref{(1)} is to simultaneously estimate multiple 3D rotations $\mathbf{R}$ and the inlier correspondence set $\mathcal{C}$ from point sets $\mathcal{Q}$ and $\mathcal{P}$.

In real applications, it becomes nearly impossible to obtain perfect observations for calculating precise rotations, as the inputs invariably contain noise and outliers. Therefore, designing robust algorithms that can produce satisfactory rotation estimates even with corrupted inputs is crucial. Unfortunately, current algorithms often face a trade-off between the accuracy of finding correspondences and the efficiency of handling large-scale data. Specifically, in the autonomous driving field, poor accuracy and low efficiency in rotation estimation can lead to serious traffic accidents, threatening people's safety. Therefore, efficiently seeking a globally optimal rotation estimate is challenging, and achieving multiple satisfactory rotation estimates simultaneously is even more difficult.

Our primary goal is to propose a rotation estimation method, which is not limited to effectively obtaining a single optimal robust rotation estimate but also can generate multiple robust rotation estimates in complex scenarios simultaneously. Specifically, this method can accurately capture multiple satisfactory rotational motions even when inputs are severely compromised by noise and outliers (i.e., mismatched points).
\subsection{Contributions}
This paper explores an accelerating outlier-robust rotation estimation method by stereographic projection. This approach addresses the challenge of accurately and efficiently estimating rotations in the presence of noise and outliers. The main contributions of this paper are summarized as follows:
\begin{itemize}
    \item The rotation estimation problem is decoupled into two subproblems, i.e., rotation axis and angle, by investigating a special geometric constraint. In other words, solving the rotation estimation problem with three degrees of freedom is transformed into solving the rotation axis with two degrees of freedom and the rotation angle with one degree of freedom. Consequently, the efficiency of solving the rotation estimation problem is significantly improved.
    \item By considering the rotation geometric constraint, solving the rotation axis becomes finding the points of maximum intersection of circles on the unit sphere. Innovatively, we use stereographic projection~\cite{wilkins2017mobius} to map the circles from a three-dimensional sphere onto a two-dimensional plane. In this way, computations in redundant spaces can be avoided, therefore increasing the efficiency of the solution.
    \item To robustly and efficiently solve the rotation axes, we introduce a spatial voting strategy to find points of maximum intersection of circles on the 2D plane. Using this strategy, we can find the optimal rotation axis and multiple rotation axes simultaneously.   
\end{itemize}
\section{Related work}
Solving the optimal rotation from data contaminated with outliers and noise is non-trivial, due to the inherently non-linear~\cite{chin2022maximum} and non-convex\cite{brynte2022tightness} nature of the \(\mathbb{SO}(3)\) space. These characteristics make rotation estimation a complex optimization challenge~\cite{bustos2016robust}, typically rendering the solution process is \(\mathcal{NP}\)-hard~\cite{peng2022arcs}. Specifically, solving the optimal rotation problem involves two main requirements: (1) robustness, and (2) efficiency. Recently, many researchers have proposed various methods, including optimization-based algorithms~\cite{maken2019speeding}, heuristic-based algorithms~\cite{nuchter2005heuristic}, deep learning-based algorithms~\cite{2021Deep}, and multi-model fitting algorithms for multiple rotation estimation~\cite{kluger2020consac}.
\subsection{Optimization-based algorithms}
Optimization-based algorithms depend on the gradient of the objective function. These methods can achieve high accuracy when initialized with suitable values. Despite their high computational efficiency, these algorithms may converge to local optima due to non-convexity. For example, in the ICP (Iterative Closest Point) algorithm~\cite{1992A}, the solution is iteratively optimized until convergence. If the initial solution is well-chosen, convergence requires only a few dozen iterations, with precision reaching \(<10^{-10}\)~\cite{wright2006numerical}. However, if the initial values are poorly chosen, these algorithms may converge to a local optimum after several dozen iterations. Consequently, they frequently serve as the concluding optimization step in various algorithmic frameworks~\cite{wright2006numerical}.
\subsection{Heuristic-based algorithms}
Heuristic-based algorithms skillfully avoid local optima and are non-deterministic in the sense that they produce a reasonable result only with a certain probability. For example, the RANSAC (RANdom SAmple Consensus) algorithm~\cite{fischler1981random} provides a common generate-and-verify pipeline for robustly removing outliers. Similarly, the FGR (Fast Global Registration) algorithm~\cite{2016Fast} uses the Geman-McClure cost function and estimates the model through progressive non-convex optimization. However, these algorithms lack guarantees for finding the optimal solution.
\subsection{Deep-learning based algorithms}
Deep-learning models can improve their robustness to noise and outliers through training on large data, maintaining high estimation accuracy even with imperfect data. Additionally, deep learning frameworks and hardware acceleration enable efficient processing of large-scale data, enhancing the efficiency of rotation estimation. however deep learning-based algorithms require training large amounts of data in advance. For instance, DGR~\cite{2020Deep} employs full convolution techniques to enhance global context capture. CN-Net~\cite{yi2018learning} combines correspondence classification with model estimation. Building on CN-Net, 3DRegNet~\cite{20193DRegNet} extends its application to 3D by integrating a regression module for handling rigid transformations. PointDSC~\cite{2021PointDSC} focuses on accelerating model generation and selection by introducing a non-local module based on spatial consistency along with neural spectral matching. Additionally, DetarNet~\cite{chen2022detarnet} offers solutions for decoupling translation and rotation. Lastly, DHVR~\cite{2021Deep} leverages deep Hough voting~\cite{qi2019deep} to extract consensus from Hough space, accurately predicting transformations.
\subsection{Multi-model fitting algorithms for multiple rotation estimation}
Classical multi-model fitting methods, such as Sequential RANSAC~\cite{zuliani2005multiransac}, proceed sequentially. This approach applies RANSAC to remove inliers of the chosen and repeats the process until a stopping criterion is reached, fitting multiple models in sequence. Although RANSAC itself is robust to outliers, the sequential processing can cause outliers from previous models to affect subsequent model estimations. Inaccurate initial model estimations can lead to error accumulation in successive models. Modern state-of-the-art methods address the multi-model fitting problem simultaneously by assigning data points to models or outlier classes using clustering~\cite{magri2014t} or optimization techniques~\cite{barath2019progressive}. However, these methods can face high computational complexity with large datasets, especially in scenarios involving multiple rotations, which require analyzing and processing numerous preference pairs. Furthermore, to the best of our knowledge, there is no such algorithm that can solve multiple rotation estimation problems efficiently. 
\begin{algorithm}[t]
    \DontPrintSemicolon
    \KwIn{Observations $\left \{ \bm{\mathit{x_{i} }} ,\bm{\mathit{y_{i} }}\right \}_{i=1}^{N}$, noise level}
    \KwOut{Optimal rotation $\mathbf{R} $}
    \BlankLine
    
    Initialize a two-dimensional accumulator $\mathbb{A}$\;
    $\theta = \left \{ \theta _{j}  \right \} _{j=1}^{J}$ by discretizing $\left [ -\pi ,\pi  \right ]$  uniformly\;

    \For{$i\in \left [ I \right ]$ and $j\in \left [ J \right ]$}
    {
    
    $\bm{\mathit{z}}_{i} = \bm{\mathit{y}}_{i}-\bm{\mathit{x}}_{i} = \left ( \bm{\mathit{a}}_{i} , \bm{\mathit{b }}_{i},\bm{\mathit{c}}_{i} \right )$ and $\bm{\mathit{a}}_{i}^{2}+\bm{\mathit{b}}_{i}^{2}+\bm{\mathit{c}}_{i}^{2}=1$\;

    Calculate basis vector $\alpha _{0} $\;
    $Proj_{\bm{\mathit{z}}_{i}} \left ( \alpha _{0} \right ) =\left ( \frac{\alpha _{0}\cdot \bm{\mathit{z}}_{i}}{\left \| \bm{\bm{\mathit{z}}_{i}} \right \|^{2}}\right )\cdot \bm{\mathit{z}}_{i}$\;

    $\bm{\alpha}_{2}=\bm{\alpha}_{1}\times\bm{\mathit{z}}_{i}$\;

    Points3d = $\bm{\alpha}_{1}\cos\theta _{j} + \bm{\alpha}_{2}\cos\theta _{j}$\;
    Points2d$\xleftarrow{\text{stereographic projection}}$points3d\;
    
    $\mathbb{A} \longleftarrow \mathbb{A} + 1$ at points2d\;

    The center of the block with the highest value in $\mathbb{A}$ is back-projected onto the sphere, yielding a point that represents the estimated rotation axis\;

    \For{$i\in \left [ I \right ]$}
    {
    $\bm{\beta}_{i} = \bm{axis}_{i}\times  \bm{\mathit{x}}_{i}$\;
    $\bm{\gamma}_{i} = \bm{axis}_{i}\times \bm{\mathit{y}}_{i}$\;
    $ angle_{i} = \arccos \frac{\bm{\beta}_{i} \cdot \bm{\gamma}_{i} }{\left \| \bm{\beta}_{i} \right \|\left \| \bm{\gamma}_{i} \right \| }$\;
    
    $\mathbb{A} \longleftarrow \mathbb{A} + 1$ at $angle_{i}$\;
    The center of the block with the highest value in $\mathbb{A}$ is returned as the estimated rotation angle\;
    } 
    }
    The rotation $\mathbf{R} $ is obtained using Rodrigues' formula.

    \caption{AORESP: Accelerating Outlier-robust Rotation Estimation by
Stereographic Projection}

\end{algorithm}
\section{Method}
This paper proposes a solution to meet two requirements, i.e., robustness and efficiency, in the robust rotation estimation problems. For efficiency, we use the stereographic projection to transform the problem of finding the maximum intersection of circles in three dimensions into a voting problem in the two-dimensional plane. For robustness, we use a spatial voting strategy to separate inliers from outliers. The idea is simple: the value with the most votes is considered inliers. This helps reduce the impact of outliers and improves the accuracy of rotation estimation.
\begin{figure}[t]
\centering
\includegraphics[width=0.5\linewidth]{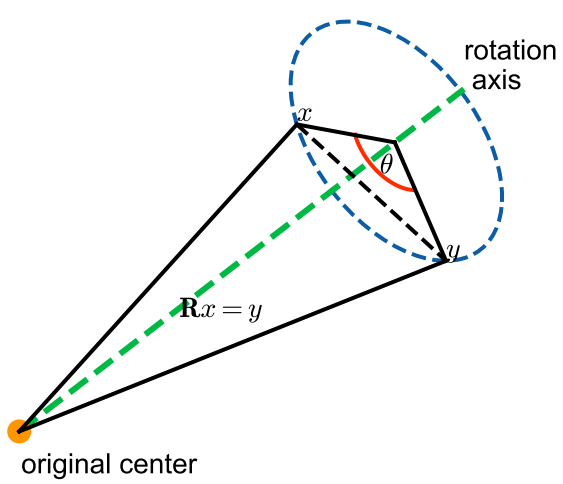}
\caption{Diagram of rotation motion, illustrating that any rotation can be represented by the rotation axis and rotation angle.}
\label{fig1}
\end{figure}
\subsection{Decoupling rotation motion}
To robustly find the optimal solution for the objective function based on the maximum consensus set, many algorithms spend a significantly longer time. For example, the Branch and Bound algorithm~\cite{liu2022globallyphd} needs to traverse the entire rotation space. Due to the 3-dimensional nature of the rotation space, this traversal causes the well-known \textit{curse of dimensionality}, severely affecting the algorithm's execution speed. To accelerate robust rotation estimation, we use spatial decoupling to simplify the complex problem while maintaining the algorithm's robustness. Specifically, this involves decoupling the rotation \(\mathbf{R}\in \mathbb{SO}(3)\)  into rotation axis \(\bm{r}\in\mathbb{S}^{2}\) and rotation angle \(\theta \in [0,\pi]\), which is known as the axis-angle representation. This process can be described by Rodrigues' rotation formula
\begin{equation}
\mathbf{R} =\exp(\theta[\bm{\mathit{r}}]_{\times})=\mathbf{I} +\sin\theta[\bm{\mathit{\bm{\mathit{r}}}}]_{\times}+(1-\cos\theta)[\bm{\mathit{r}}]_{\times}^{2},
\end{equation}
where $[\mathit{\bm{\mathit{r}}}]_{\times}$is a skew-symmetric matrix for cross-product.

Following the principles of rotational kinematics (refer to Fig.\ref{fig1}), we establish a constraint-based solely on the rotation axis (shown as Equations \ref{8}):
\begin{align}
&\quad\mathbf{R}\bm{\mathit{x}} = \bm{\mathit{y}} \label{5} 
\Rightarrow \bm{\mathit{r^{T} }}\mathbf{R}\bm{\mathit{x}} = \bm{\mathit{r^{T} }}\bm{\mathit{y}} \\
&\Rightarrow \bm{\mathit{r^{T} }}\bm{\mathit{x}} = \bm{\mathit{r^{T} }}\bm{\mathit{y}} 
\Rightarrow \bm{\mathit{r^{T} }}(\bm{\mathit{x}}-\bm{\mathit{y}}) = 0
\label{8}
\end{align}
Therefore, we can reconstruct the objective function using the maximum consensus set (shown as Equation \ref{9}) based on the constraints of the rotation axis.
\begin{equation}
 \sum_{i=1}^{N}\mathbb{I}(\left \| \bm{\mathit{r^{T}}} (\bm{\mathit{x}}_{i}-\bm{\mathit{y}}_{i})  \right \|< \varepsilon)
 \label{9}
\end{equation}

The rotation axis \(\bm{r}\) is a unit vector representing the direction of the rotation axis. Geometrically, \(\bm{r}\) is a point on the surface of the unit sphere, i.e., \(\bm{r} \in \mathbb{S}^{2}\). The goal is to find the optimal rotation axis on the unit sphere. Once the optimal axis is identified, calculating the rotation angle is straightforward. In summary, the challenge of estimating the optimal rotation (initially involving three degrees of freedom) is divided into identifying the optimal axis (a two degree of freedom problem) and determining the optimal rotation angle (a one degree of freedom problem)
\section{Stereographic Projection and Voting Strategies}
\begin{definition}[stereographic projection~\cite{wilkins2017mobius}]
    In the field of geometry, stereographic projection is a mapping that projects a sphere onto a plane. Except for the projection point, the stereographic projection is smooth, bijective, and conformal across the entire sphere. Conformality means that 3D circles on the sphere are projected as 2D circles on the plane. However, this mapping does not preserve distances or areas.
\end{definition}
After decoupling the optimal rotation estimation problem, we focus on robustly and efficiently determining the optimal rotation axis. To achieve this, we implement the stereographic projection strategy. This strategy transforms the spherical optimization problem into a planar one, enabling an \(\mathcal{O(N)}\) optimization algorithm, and ensuring efficient and robust axis estimation.

Specifically, according to the geometric constraints of rotational motion, \({\mathbf{R}\bm{\mathit{x}}} = \bm{\mathit{y}} \Rightarrow \bm{\mathit{r^{T}}} (\bm{\mathit{x}}-\bm{\mathit{y}})=0\), which geometrically indicates that \(\bm{\mathit{r}}\) is perpendicular to \(\bm{\mathit{x}}-\bm{\mathit{y}}\), i.e,  \(\angle (\bm{\mathit{r}}
,\bm{\mathit{x}}-\bm{\mathit{y}})=90^{\circ}\). By normalizing all lengths to define \(\bm{\mathit{z}}=\frac{\bm{\mathit{x}}-\bm{\mathit{y}} }{\left \| \bm{\mathit{x}}-\bm{\mathit{y}} \right \| }\), we establish \(\bm{\mathit{z}}^{T}\bm{\mathit{r}}=0\). With $\bm{N}$ input sets, we derive a series of linear equations:
\begin{equation}
\begin{bmatrix}
 \bm{\mathit{z}}_{1}^T, 
 \bm{\mathit{z}}_{2}^T,
 \cdots,
 \bm{\mathit{z}}_{N}^T
\end{bmatrix}\bm{\mathit{r}}=\begin{bmatrix}
 0, 
  0,
 \cdots,
  0
\end{bmatrix} \label{(10)}
\end{equation}
In Equation \ref{(10)}, when \(\bm{N}\) is greater than or equal to 2, the rotation axis can be determined in the least squares sense~\cite{viklands2006algorithms}. However, if there are outliers or errors in the input, the least squares solution will significantly deviate from the correct solution. Geometrically, given the $i$-th \(\bm{\mathit{z}}_{i}\) which is a vector point on the sphere (see Fig.\ref{fig2}), the rotation axis that satisfies the constraint must be perpendicular to \(\bm{\mathit{z}}_{i}\). Thus, the rotation axis that satisfies the constraint must lie in a 3D circle. If multiple \(\bm{\mathit{z}}_{i}\) are given, multiple circles will intersect at a point, which indicates the direction of the rotation axis. Since finding the intersection points of 3D circles on a sphere is time-consuming, we use stereographic projection to transform the problem into finding intersection points on a 2D plane (see Fig.\ref{fig2}). This method is particularly useful for solving the optimal rotation axis problem. 

Specifically, for each point on the sphere with coordinates \((\mathit{a}, \mathit{b}, \mathit{c})\), there is a corresponding point \((\mathit{A}, \mathit{B})\) on the plane, defined as follows\cite{Li:Math2220Notes}
\begin{equation}
 (\mathit{A}, \mathit{B}) = \left(\frac{\mathit{a}}{1-\mathit{c}}, \frac{\mathit{b}}{1-\mathit{c}}\right)
\end{equation}
The inverse projection formula is given by
\begin{multline}
(\mathit{a},\mathit{b},\mathit{c})=\\
(\frac{2\mathit{A}}{1+\mathit{A}^{2}+\mathit{B}^{2}},\frac{2\mathit{B}}{1+\mathit{A}^{2}+\mathit{B}^{2}},\frac{-1+\mathit{A}^{2}+\mathit{B}^{2}}{1+\mathit{A}^{2}+\mathit{B}^{2}})
\label{12}
\end{multline}
Any circle on the unit sphere can be
written as the intersection an plane in $\mathbb{S}^{2}$ and the sphere
\begin{equation}
\begin{cases}
  X^2 + Y^2 + Z^2 = 1, \\
  AX + BY + CZ + D = 0.
\end{cases}
\end{equation}
Observe,
\begin{align}
(\mathit{x}+\frac{\mathit{A}}{\mathit{C}+\mathit{D}})^{2} + (\mathit{y}+\frac{\mathit{B}}{\mathit{C}+\mathit{D}})^{2} &= \notag
\\
&\frac{\mathit{A}^{2}+\mathit{B}^{2}}{(\mathit{C}+\mathit{D})^{2}}+\frac{\mathit{C}-\mathit{D}}{\mathit{C}+\mathit{D}} 
\label{13}
\end{align}
Equation \ref{13} represents a circle in geometry.
\begin{table*}
\setlength{\tabcolsep}{2pt}
\caption{Success rates of methods run on the scene pairs of the 3DMatch dataset for which the ground-truth rotation and translation are provided (where rotation error smaller than 10 degrees means a success)}
\centering
\begin{tabularx}{\textwidth}{@{}l*{9}{X}@{}}
\toprule
Scene Type & Kitchen & Home 1 & Home 2 & Hotel 1 & Hotel 2 & Hotel 3 & Study Room & MIT Lab & Overall \\
\# Scene Pairs & 506     & 156    & 208    & 226     & 104     & 54      & 292        & 77      & 1623    \\ \midrule
TEASER++~\cite{2021TEASER}      & 99.0\%   & 98.1\%  & 94.7\%  & 98.7\%    & 99.0\%   & 98.1\%   & 97.0\%      & 94.8\%   & 97.72\%  \\
$\textnormal{ARCS+}\textsubscript{OR}$~\cite{peng2022arcs}      & 98.4\%   & 97.4\%  & 95.7\%  & 98.7\%   & 98.1\%   & 100.0\%  & 97.3\%      & 96.1\%   & 97.72\%  \\
\hline
Ours          & 100.0\%  & 100.0\% & 100.0\% & 100.0\%  & 100.0\%  & 100.0\%  & 100.0\%     & 100.0\%  & 100.0\%  \\ \bottomrule
\end{tabularx}
\label{table3}
\end{table*}
\begin{figure*}
\centering
\begin{tabular}{cccccc}
\includegraphics[width=0.147\linewidth]{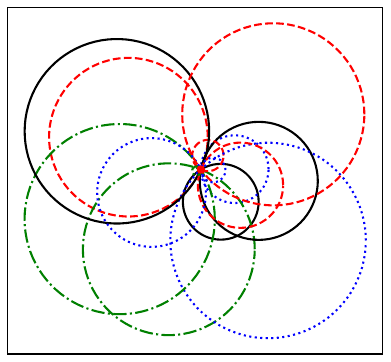}
&\includegraphics[width=0.152\linewidth]{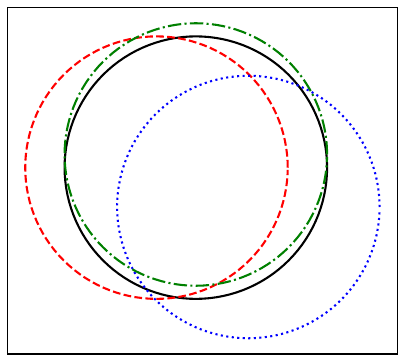}
&\includegraphics[width=0.132\linewidth]{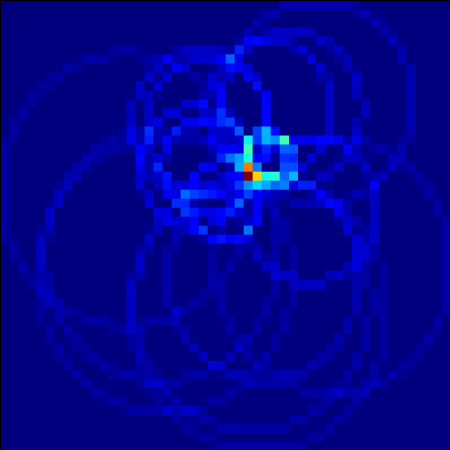}
&\includegraphics[width=0.132\linewidth]{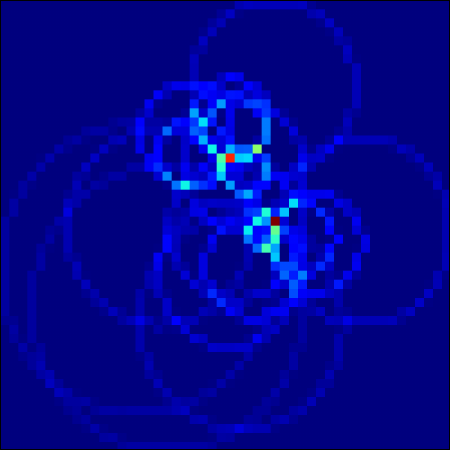}
&\includegraphics[width=0.158\linewidth]{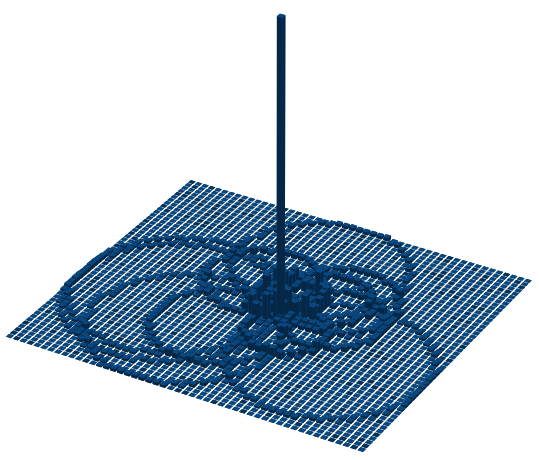}
&\includegraphics[width=0.16\linewidth]{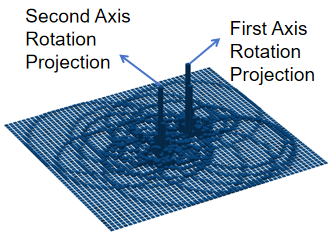}
\\(a)&(b)&(c)&(d)&(e)&(f)
\end{tabular}
\caption{(a) With the perfect inputs, many project circles are across the same point, corresponding to the rotation axis. (b) With incorrect inputs, there are multiple wrong intersections, which will seriously affect the results of the pose estimation if not handled properly. (c) The number in each cell represents the number of circles that pass through the cell. (d) There are two rotation axes in this setting. (e) The 3D visualization histogram, with the highest peak being the position of the desired optimal rotation axis. (f) The two-dimensional projections of the two rotational axes are obtained from the histogram.}
\label{fig3}
\end{figure*}
\subsection{Optimal rotation axis calculation}
After stereographic projection, circles are mapped onto a plane. This transforms the task into finding a point in $\mathbb{R}^{2}$ intersected by the most circles. This point indicates the direction of the optimal rotation axis. To robustly determine this optimal point, discretization, and voting strategies are employed. Specifically, the solution domain for the rotation axis is discretized into a 2D accumulator. Theoretically, the cell containing the optimal rotation axis will have the most circles passing through it, as illustrated in Fig.\ref{fig3}. The axis's direction in three-dimensional space is inferred by reversing the stereographic projection.
\subsection{Optimal rotation angle calculation}
After determining the optimal rotation axis, estimating the rotation angle becomes the next task in the rotation estimation problem. This step is accomplished by applying Rodrigues' rotation formula.
\begin{equation}
\left(\mathbf{I} +\sin\theta[\mathit{\bm{\mathit{r}}}]_{\times}+(1-\cos\theta)[\mathit{\bm{\mathit{r}}}]_{\times}^{2}\right)\bm{\mathit{x}} =\bm{\mathit{y}}
\label{15}
\end{equation}
given $\bm{r}$, Equation \ref{15} results in a trigonometric equation related to \(\theta\), allowing for the determination of the optimal rotation angle. Importantly, each input pair \(\{\bm{\mathit{x}}_{i}, \bm{\mathit{y}}_{i} \}\) produces a specific rotation angle \(\theta _{i}\). Therefore, the optimal rotation angle is determined through histogram voting~\cite{xing2024efficient}.
\subsection{Rotation recovery} At this point, we have determined the optimal rotation axis and angle. Then we use Rodrigues' rotation formula to reconstruct the rotation. In other words, this method replaces the complex three degrees of freedom optimal rotation estimation problem with two simpler problems: one with two degrees of freedom and one with one degree of freedom. This approach improves both efficiency and robustness.
\begin{table}[t]
\centering
\caption{Quantitative results on 3DMatch dataset. Methods with * are correspondence-free methods.}
\begin{tabular}{l|ll|ll}
\hline
\multicolumn{1}{c|}{} & \multicolumn{2}{c|}{FPFH}                        & \multicolumn{2}{c}{FCGF}                        \\
                      & \multicolumn{2}{l|}{(traditional descriptor)} & \multicolumn{2}{l}{(learning descriptor)} \\
                      & RE(°)                   & Times(sec)                  & RE(°)                   & Times(sec)                 \\ \hline
DCP*~\cite{2019Deep}                  & 8.42                    & 0.07                   & 8.42                    & 0.07                  \\
PointNetLK*~\cite{2019PointNetLK}          & 8.04                    & 0.12                   & 8.04                    & 0.12                  \\
3DRegNet~\cite{20193DRegNet}             & 3.75                    & 0.05                   & 2.74                    & 0.05                  \\
DGR~\cite{2020Deep}                  & 2.45                    & 1.53                   & 2.28                    & 1.53                  \\
DHVR~\cite{2021Deep}                  & 2.78                    & 3.92                   & 2.25                    & 3.92                  \\
DHVR-Origin~\cite{2021Deep}          & 2.06                    & 0.46                   & 2.08                    & 0.46                  \\
PointDSC~\cite{2021PointDSC}            & 2.03                    & 0.10                   & 2.08                    & 0.10                  \\
SM~\cite{Leordeanu2005A}                    & 2.94                    & 0.03                   & 2.29                    & 0.03                  \\
ICP*~\cite{1992A}                  & 7.93                    & 0.25                   & 7.93                    & 0.25                  \\
FGR~\cite{2016Fast}                   & 4.96                    & 0.89                   & 2.90                    & 0.89                  \\
GC-RANSAC~\cite{8578802}             & 2.33                    & 0.55                   & 2.33                    & 0.55                  \\
RANSAC-1M~\cite{fischler1981random}             & 4.05                    & 0.97                   & 3.05                    & 0.97                  \\
RANSAC-2M~\cite{fischler1981random}& 4.07                    & 1.63                   & 2.71                    & 1.63                  \\
RANSAC-3M~\cite{fischler1981random}            & 3.95                    & 2.86                   & 2.69                    & 2.86                  \\
CG-SAC~\cite{2020Compatibility}                & 2.40                    & 0.27                   & 2.42                    & 0.27                  \\
$\text{SC}^{2}$-\text{PCR}~\cite{9878510}               & 2.18                    & 0.11                   & 2.08                    & 0.11                  \\
Ours                  & \textbf{1.05 }                   & \textbf{0.01  }                 & \textbf{1.10 }                   & \textbf{0.01 }                 \\ \hline
\end{tabular}
\label{table1}
\end{table}
\begin{table}[h]
\caption{Quantitative results on KITTI dataset}
\centering
\begin{tabular}{l|ll}
\hline
\multicolumn{1}{c|}{} & \multicolumn{2}{c}{FPFH}                           \\
                      & \multicolumn{2}{l}{(traditional based descriptor)} \\
                      & RE(°)                    & Times(sec)                   \\ \hline
DHVR~\cite{2021Deep}                  & -                        & 0.83                    \\
DGR~\cite{2020Deep}                   & 1.64                     & 2.29                    \\
PointDSC~\cite{2021PointDSC}              & 0.35                     & 0.45                    \\
FGR~\cite{2016Fast}                   & 0.86                     & 3.88                    \\
RANSAC~\cite{fischler1981random}               & 1.55                     & 5.43                    \\
CG-SAC~\cite{2020Compatibility}                & 0.73                     & 0.73                    \\
SC2-PCR~\cite{9878510}               & 0.32                     & 0.31                    \\
Ours                  & \textbf{0.14}                     & \textbf{0.01 }                   \\ \hline
\end{tabular}
\label{table2}
\end{table}
\begin{figure*}[t]
\centering
\includegraphics[width=1\linewidth]{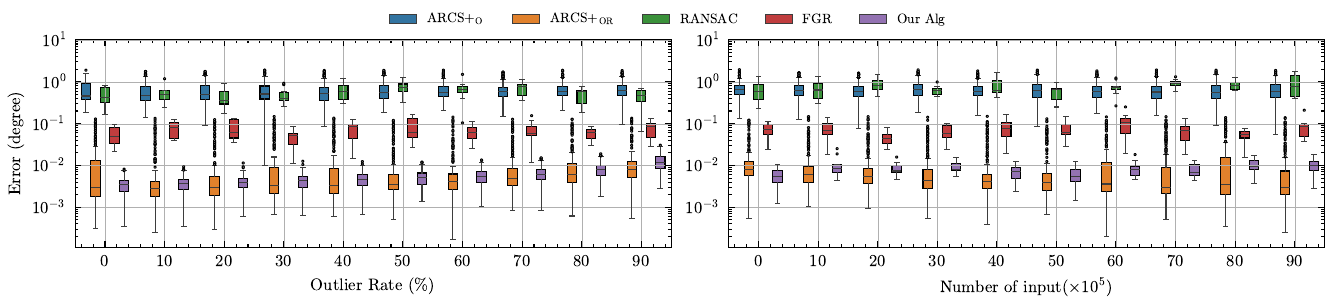}
\caption{Comparative experiments. The left calculates the rotation errors using inputs with different outlier rates when the total number is $10^{5}$. The right calculates the rotation errors using different amounts of inputs when the outlier rate is 90\%.}
\label{fig5}
\end{figure*}
\begin{figure*}[t]
\centering
\includegraphics[width=1\linewidth]{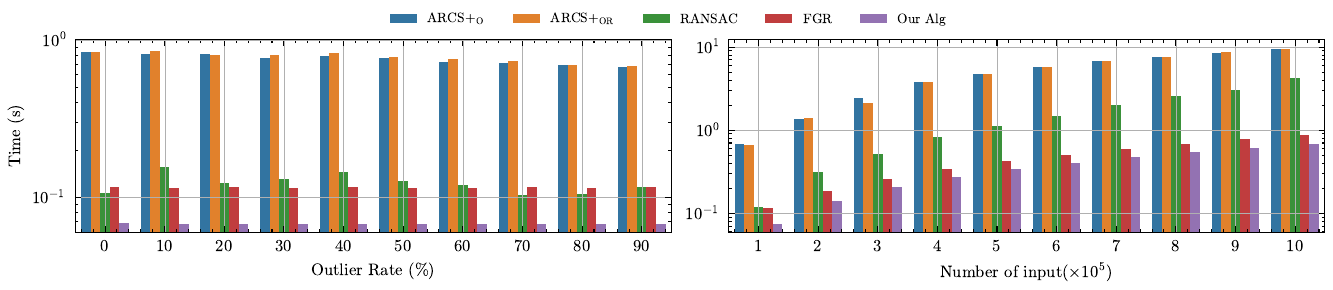}
\caption{Comparative experiments. The left calculates the running time using inputs with different outlier rates when the total number is $10^{5}$. The right calculates the running time using different amounts of inputs when the outlier rate is 90\%.}
\label{fig6}
\end{figure*}
\begin{figure*}[t]
\centering
\includegraphics[width=1\linewidth]{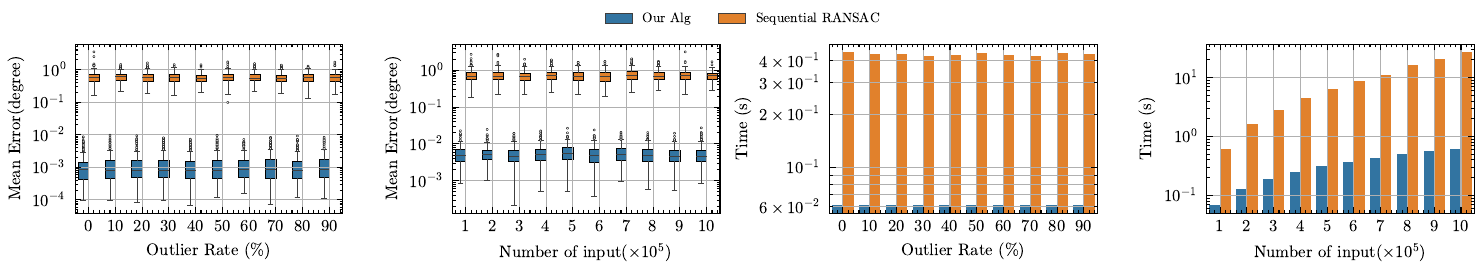}
\caption{Comparative experiments between our algorithm and Sequential RANSAC for multiple rotation estimations. The first and third calculate the average rotation errors and running time using inputs with different outlier rates when the total number is $10^{5}$. The second and fourth calculate the average rotation errors and running time using different amounts of input data when the outlier rate is 90\%.}
\label{fig8}
\end{figure*}
\begin{figure}[t]
\centering
\includegraphics[width=1\linewidth]{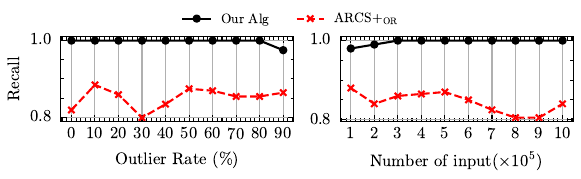}
\caption{A comparative experiment is conducted between $\textnormal{ARCS+}\textsubscript{OR}$ and Our Algorithm (Calculating the proportion of values with rotation errors which less than 0.25 out of 200 experiments. higher is better).}
\label{fig7}
\end{figure}
\subsection{Multi-model estimation rotation }
In fact, utilizing a spatial voting strategy allows not only for the identification of the optimal rotation axis but also for the simultaneous detection of all rotation axes (as shown in fig.\ref{fig3}). Each rotation axis corresponds to a specific rotation angle. A rotation can be derived for every pair of axes and angles, enabling us to obtain multiple robust rotation estimations simultaneously.
\begin{figure}[t]
\centering
\includegraphics[width=1\linewidth]{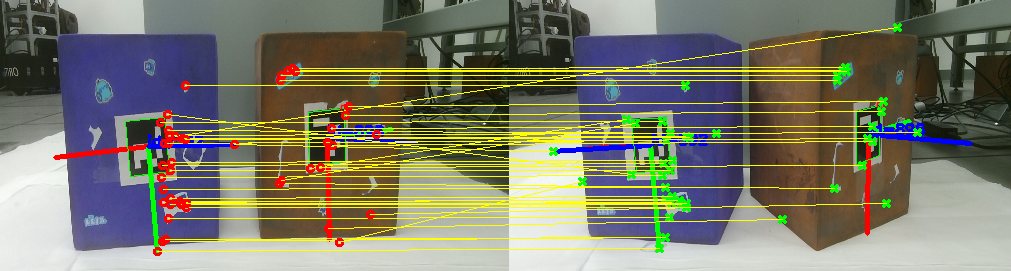}
\caption{An example from our collected data. Red circles (o) represent feature matching points in the image before rotation. Green cross ($\times$) signs represent feature matching points in the image after rotation. Yellow lines indicate the correspondences. (For better visualization, we select only a few matching points to display.)}
\label{fig10}
\end{figure}
\begin{table}[t]
\caption{Results of the multi-model experiment using our collected data (the rotation error in the table is the average of the errors of two rotations.)}
\centering
\begin{tabular}{@{}cccc@{}}
\toprule
Scenes  & Feature Number & Avg RE(°) & Times(sec) \\ \midrule
1  & 295            & 0.212        & 0.01      \\
2  & 306            & 0.237        & 0.01      \\
3  & 299            & 0.206        & 0.01      \\
4  & 314            & 0.175        & 0.01      \\
5  & 301            & 0.198        & 0.01      \\
6  & 281            & 0.254        & 0.01      \\
7  & 326            & 0.173        & 0.01      \\
8  & 286            & 0.203        & 0.01      \\
9  & 311            & 0.191        & 0.01      \\
10 & 303            & 0.188        & 0.01      \\ \bottomrule
\end{tabular}
\label{table4}
\end{table}
\section{EXPERIMENTS}
To evaluate the accuracy and robustness of the proposed method, comparative experiments are systematically conducted using controlled synthetic and real-world data. 
All experiments are validated on a PC with 16G RAM and GeForce RTX 4080. Unless otherwise specified, each synthetic experiment was conducted 200 times.

\subsection{Comparative experiments on robust rotation estimation
}
\subsubsection{Synthetic data experiments}
First, we randomly generate \( N \) synthetic three-dimensional point clouds and normalize these data to generate unit-length vectors $\left \{ \bm{\mathit{x}}_{i}  \right \} _{i=1}^{N} $. A randomly generated rigid rotation  \(\mathbb{SO}(3)\) is applied to the normalized data to generate $\left \{ \bm{\mathit{y}}_{i}  \right \} _{i=1}^{N}$. To simulate real-world outliers and noise, random points are added to mimic outliers with a ratio of \( \rho=\frac{N_{outlier}}{N} \), and noise is introduced with a standard deviation of \( \sigma=0.01\). The rotation error \( e_{rot} \) is defined by
\begin{equation}
 e_{rot} = \arccos\left(0.5\left({\text{Tr}(R_{gt}^T R_{opt}) - 1}\right) \right)
\end{equation}
where \(\text{Tr}(\cdot)\) denotes the trace function of a square matrix, \( R_{gt} \) is the ground truth, and \( R_{opt} \) is the optimal rotation computed by the proposed algorithm. 

\textbf{Results.}
In this experiment, we compare our algorithm with $\textnormal{ARCS+}\textsubscript{O}$~\cite{peng2022arcs}, $\textnormal{ARCS+}\textsubscript{OR}$~\cite{peng2022arcs}, RANSAC~\cite{fischler1981random}, FGR~\cite{2016Fast}, GORE~\cite{bustos2017guaranteed}, and TEASER++~\cite{2021TEASER}. Notably, TEASER++ encounters memory issues, and GORE's running time exceeds 12 hours.
Regarding accuracy, as shown in Figure \ref{fig5}, all algorithms perform well, but only $\textnormal{ARCS+}\textsubscript{OR}$ matches our algorithm. The other algorithms are less accurate. To further demonstrate the superiority of our algorithm, we compare the recall of $\textnormal{ARCS+}\textsubscript{OR}$ and our algorithm, as depicted in Figure \ref{fig7}. Higher recall indicates better accuracy. Thus our algorithm outperforms all others in accuracy. In terms of efficiency, as illustrated in Figure \ref{fig6}, our algorithm is significantly faster than all other algorithms.
\subsubsection{Real-world data experiments}

We use two distinct datasets to demonstrate our algorithm's performance in both indoor and outdoor environments: 3DMatch~\cite{zeng20173dmatch} and KITTI~\cite{geiger2012we}. The 3DMatch benchmark includes over 1,000 point clouds from eight different indoor scenes, such as kitchens and hotels, each containing between 77 to 506 point clouds with over 100,000 points each. For outdoor scenes, we select the KITTI dataset, choosing 8 to 10 scenes and securing 555-point cloud pairs for testing. We downsample point clouds using voxel grids—5 cm for 3DMatch and 30 cm for KITTI. We then extract local feature descriptors and perform matches to establish hypothetical correspondences. We use traditional descriptors like Fast Point Feature Histograms (FPFH)~\cite{rusu2009fast} and learned descriptors like Fully Convolutional Geometric Features (FCGF)~\cite{choy2019fully}. These processed point clouds validate our algorithm's effectiveness across varied settings. Since our algorithm focuses on rotation estimation, we use ground truth for translation.

\textbf{Results.} First, we present the results on the 3DMatch dataset~\cite{zeng20173dmatch} in Table \ref{table1}. We compare our method with 13 baselines: DCP~\cite{2019Deep}, PointNetLK~\cite{2019PointNetLK}, 3DRegNet~\cite{20193DRegNet}, DGR~\cite{2020Deep}, DHVR~\cite{2021Deep}, PointDSC~\cite{2021PointDSC}, SM~\cite{Leordeanu2005A}, ICP~\cite{1992A}, FGR~\cite{2016Fast}, 
GC-RANSAC~\cite{8578802}, RANSAC~\cite{fischler1981random}, CG-SAC~\cite{2020Compatibility} and $\text{SC}^{2}$-\text{PCR}~\cite{9878510}. The first six methods are deep learning-based, while the last seven are traditional. For robustness, when using the FPFH descriptor, PointDSC is the most robust algorithm after ours, with our algorithm improving the rotation error by 48\% (as shown in Table \ref{table1}). When using the FCGF descriptor, the most robust algorithms after ours are DHVR-Origin, PointDSC, and $\text{SC}^{2}$-\text{PCR}, with our algorithm improving the rotation error by 47\% (as shown in Table \ref{table1}). For efficiency, whether using FPFH or FCGF, SM is the fastest algorithm after ours, but our algorithm runs three times faster than SM (as shown in Table \ref{table1}).

Second, we present the results on the outdoor KITTI dataset in Table \ref{table2}. We compare the performance of DHVR, DGR, PointDSC, RANSAC, FGR, CG-SAC, and $\text{SC}^{2}$-\text{PCR}. DHVR, DGR, and PointDSC are deep learning-based methods, while RANSAC, FGR, and CG-SAC are traditional non-learning methods. For robustness and efficiency, aside from our algorithm, $\text{SC}^{2}$-\text{PCR} performs the best. However, our algorithm improves robustness by 56\% and speed by 30\% compared to $\text{SC}^{2}$-\text{PCR}.
\subsection{Comparative experiments on multiple rotations
}
\subsubsection{Synthetic data experiments}
We use the same data setup and rotation error calculation method as previous experiments. The only difference is that there are multiple rotations to be estimated. 

\textbf{Results.} As shown in Figure \ref{fig8}, our algorithm excels in solving multiple rotations simultaneously compared to the Sequential RANSAC algorithm. It significantly improves robustness and efficiency. Notably, our algorithm runs nearly 100 times faster than Sequential RANSAC, highlighting its superior performance in solving multiple rotation estimation.
\subsubsection{Real-world data experiments}
To better verify the superiority of our algorithm in solving multi-rotation problems, we conduct a real-world experiment utilizing data collected from real-world scenes, as depicted in Figure \ref{fig10}. We employ the Kinect v2 depth camera as our primary tool for data gathering, capable of simultaneously capturing the spatial position information of two boxes before and after rotation. Simultaneously, we specifically utilize ArUco markers for precise spatial calibration, furnishing us with two ground truths as benchmarks for subsequent evaluation of algorithm performance. Following this, we apply the classic Scale-Invariant Feature Transform (SIFT) algorithm~\cite{wu2013comparative} to efficiently identify and match corresponding feature points in the pre-rotation and post-rotation images. Ultimately, we transform these matched feature points into 3D point cloud data, where the pre-rotation data serves as the observation points, and the post-rotation data functions as the target points.

\textbf{Results.} As shown in Table \ref{table4}, we collect a total of 10 scenes for verification. We obtain 10 rotation errors in total, with an average rotation error of 0.2037, a variance of 0.0006, and a processing time of 0.01 seconds. This indicates that our method is both fast and robust when dealing with multiple rotations in real-world applications.
\section{CONCLUSION}
This paper presents an accelerating outlier-robust rotation estimation by
stereographic projection. The proposed approach first decouples the rotation \(\mathbf{R}\in \mathbb{SO}(3)\) into  rotation axis \(\bm{r}\in\mathbb{S}^{2}\) and rotation angle \(\theta\in[0,\pi]\), establishing constraints solely on the rotation axis. Then, it employs stereographic projection and spatial voting to identify the optimal rotation axis, followed by determining the optimal rotation angle. Finally, the Rodrigues’
rotation formula is used to derive the optimal rotation. Additionally, the method can simultaneously identify multiple rotation axes, resulting in multiple rotations. Extensive experiments demonstrate that our method achieves high efficiency and accuracy.

\bibliographystyle{IEEEtran}
\bibliography{main}
\end{document}